\documentclass[preprint,12pt]{elsarticle}

\usepackage{amsmath,amssymb,amsfonts}
\usepackage{algorithmic}
\usepackage{bbm}
\usepackage{commath}
\usepackage{graphicx}
\usepackage{mathtools}

\usepackage{textcomp}
\usepackage{xcolor}
\def\BibTeX{{\rm B\kern-.05em{\sc i\kern-.025em b}\kern-.08em
    T\kern-.1667em\lower.7ex\hbox{E}\kern-.125emX}}
    
\usepackage{cleveref}

\usepackage{bm}
\DeclareMathAlphabet{\mathsfit}{\encodingdefault}{\sfdefault}{m}{sl}
\SetMathAlphabet{\mathsfit}{bold}{\encodingdefault}{\sfdefault}{bx}{n}

\usepackage{adjustbox}
\usepackage{tikz,forest}
\usetikzlibrary{arrows.meta}
\forestset{
  default preamble={
    before typesetting nodes={
      !r.replace by={[, coordinate, append]}
    },
    where n children=0{
      tier=word,bottom color=green!20,top color=green!20
    }{
      diamond, aspect=2,bottom color=red!10,top color=red!10
    },
    where level=0{}{
    },
    for tree={
      edge+={thick, -Latex},
      math content,
      s sep'+=1cm,
      draw,
      thick,
      edge path'={ (!u) -| (.parent)},
    }
  }
}

\journal{Digital Signal Processing}

\begin{document}

\begin{frontmatter}



\title{A Hybrid Framework for Sequential Data Prediction with End-to-End Optimization}


\author[inst1]{Mustafa E. Aydin\corref{cor1}}
\cortext[cor1]{Corresponding author.}
\ead{enesa@ee.bilkent.edu.tr}

\author[inst1]{Suleyman S. Kozat}
\ead{kozat@ee.bilkent.edu.tr}

\affiliation[inst1]{organization={Department of Electrical and Electronics Engineering},
            addressline={Bilkent University}, 
            city={Ankara},
            postcode={06800}, 
            country={Turkey}}

\begin{abstract}
We investigate nonlinear prediction in an online setting and introduce a hybrid model that effectively mitigates, via an end-to-end architecture, the need for hand-designed features and manual model selection issues of conventional nonlinear prediction/regression methods. In particular, we use recursive structures to extract features from sequential signals, while preserving the state information, i.e., the history, and boosted decision trees to produce the final output. The connection is in an end-to-end fashion and we jointly optimize the whole architecture using stochastic gradient descent, for which we also provide the backward pass update equations. In particular, we employ a recurrent neural network (LSTM) for adaptive feature extraction from sequential data and a gradient boosting machinery (soft GBDT) for effective supervised regression. Our framework is generic so that one can use other deep learning architectures for feature extraction (such as RNNs and GRUs) and machine learning algorithms for decision making as long as they are differentiable. We demonstrate the learning behavior of our algorithm on synthetic data and the significant performance improvements over the conventional methods over various real life datasets. Furthermore, we openly share the source code of the proposed method to facilitate further research.
\end{abstract}




\end{frontmatter}


\section{Introduction}
\subsection{Background}
\label{sec:backg}
We study nonlinear prediction in a temporal setting where we receive a data sequence related to a target signal and estimate the signal's next samples. This problem is extensively studied in the signal processing and machine learning literatures since it has many applications such as in electricity demand \cite{fore_elec}, medical records \cite{fore_medic}, weather conditions \cite{fore_weather} and retail sales \cite{fore_retail}. Commonly, this problem is studied as two disjoint sub-problems where features are extracted, usually by a domain expert, and then a decision algorithm is trained over these selected features using possibly different feature selection methods. As shown in our simulations, this disjoint optimization can provide less than adequate results for different applications since the selected features may not be the best features to be used by the selected algorithm. To remedy this important problem, we introduce an algorithm which performs these two tasks jointly and optimize both the features and the algorithm in an end-to-end manner in order to minimize the final loss. In particular, we use recursive structures to extract features from sequential data, while preserving the state information, i.e., the history, and boosted decision trees to produce the final output, which are shown to be very effective in several different real life applications \cite{boosting_1,boosting_2}. This combination is then jointly optimized to minimize the final loss in an end-to-end manner.  

The existing well-known statistical models (such as autoregressive integrated moving average and exponential smoothing) for regression are robust to overfitting, have generally fewer parameters to estimate, easier to interpret due to the intuitive nature of the underlying model and amenable to automated procedures for hyperparameter selection \cite{the_forecasting_paper}. However, these models have very strong assumptions about the data such as stationarity and linearity, which prevent them from capturing nonlinear patterns that real life data tend to possess \cite{nonlinear_life}. Machine learning-based models, on the other hand, are data-driven and free of the strong statistical assumptions. Recurrent neural networks (RNN) and in particular long short-term memory neural networks (LSTM) are one of such models that excel at representing the data, especially the sequential data, in a hierarchical manner via nonlinear modules stacked on top of each other \cite{ant_yt}. Thanks to the feedback connections to preserve the history in memory \cite{rnn_itself}, they are widely used in sequential learning tasks. Nevertheless, a fully connected layer usually employed in the final layer of these hidden layers often hinders their regression ability (e.g., in \cite{rnn_nondense}, authors use an attention layer for more accurate judgments). An alternative to these deep learning models is the tree-based models that learn hierarchical relations in the data by splitting the input space and then fitting different models to each split \cite{ant_yt}. Such a data-driven splitting and model fitting are shown to be very effective in real life applications \cite{trees_fine_1, trees_fine_2}. Among such models, gradient-boosted decision trees (GBDT) \cite{gbdt_friedman} are promising models that work via sequentially combining weak learners, i.e., decision trees. Specialized GBDTs, such as XGBoost \cite{xgb_paper} and LightGBM \cite{lgb_paper}, demonstrated excellent performance at various time series prediction problems \cite{m5_comp}. Although they were not designed to process sequential data, GBDTs can be used by incorporating temporal side information, such as lagged values of the desired signal, as components of the feature vectors. However, this need for domain-specific feature engineering and independent design of the algorithm from the selected features hinder their full potential in sequential learning tasks since both processes are time consuming in most applications and require domain expertise \cite{finans_paper, cyber_paper}.

Here, we effectively combine LSTM and GBDT architectures for nonlinear sequential data prediction. Our motivation is to use an LSTM network as a feature extractor from the sequential data in the front end, mitigating the need for hand-designed features for the GBDT, and use the GBDT to perform supervised regression on these learned features while enjoying joint optimization. In particular, we connect the two models in an end-to-end fashion and jointly optimize the whole architecture using stochastic gradient descent in an online manner. To be able to learn not only the feature mapping but also the optimal partition of the feature space, we use a \emph{soft} gradient boosting decision tree (sGBDT) \cite{sgbm_paper} that employs soft decision trees \cite{irsoy_sdt,hinton_sdt} as the weak learners. We emphasize that our framework is generic so that one can use other deep learning architectures for feature extraction (such as RNNs and GRUs)  and machine learning algorithms for decision making as long as they are differentiable. 

\subsection{Related Work}
Combination of decision trees with neural networks has attracted wide interest in the machine learning literature. In \cite{ndf_paper}, authors propose a neural decision forest where they connect soft decision trees to the final layer of a CNN architecture. This combination provided the state-of-the-art performance in the ImageNet dataset \cite{image_net_dataset}. Our work differs from this method since we use boosting with fixed-depth trees instead of bagging and we jointly optimize the whole architecture whereas they employ an alternating optimization procedure. In \cite{hinton_sdt}, Frosst and Hinton first employ a neural network and then using its predictions along with the true targets, train a soft decision tree. Their main goal is to get a better understanding of the decisions a neural network makes by distilling its knowledge into a decision tree. They, however, focus on classification, use two separate training phases by design, and favor explainability over model performance. The adaptive neural trees proposed in \cite{ant_yt} uses a soft decision tree with ``split data, deepen transform and keep'' methodology where neural networks are the transformers in the edges of the tree. The tree is thereby grown step-wise and overall architecture necessitates both a ``growth'' phase and a ``refinement'' phase (where parameters are actually updated). Our architecture, however, does not embed neural networks inside the trees but places it in a sequential manner, does not grow trees but uses fixed depth weak learners and performs the optimization via a single backpropagation pass. References \cite{nrf_paper,tel_paper} also take similar approaches to the aforementioned models but none of the proposed models are designed to process sequential data, which renders them unsuitable for time series forecasting tasks. Perhaps the closest architecture to ours is \cite{disj_paper} where authors aim a forecasting task with a hybrid model of an LSTM network and XGBoost. Nevertheless, they employ a disjoint architecture in that they have a three stage training: twice for XGBoost and once for LSTM, which not only increases the computational time but also retains it from enjoying the benefits of an end-to-end optimization \cite{end_to_end_paper_1, end_to_end_paper_2}. Adaptive feature extraction-based solutions for images have been studied in \cite{stan_paper_1_anew_payl, stan_paper_2_adap_pay_feat}, where the authors develop adaptive and secure image steganographic algorithms based on input channel correlations and feature extraction, respectively. In \cite{img_forge_paper_robust_img_op}, authors propose a two-stream convolutional neural network (CNN) to detect image operator chains. Their end-to-end model is parallel in the sense that the input is fed to two different convolutional network architectures, outputs of which are merged together at the end to improve performance. Our framework is more flexible to perform sequential data regression as they focus on image classification and introduce a specific CNN architecture, however.

Unlike previous studies, our model is armed with joint optimization of a \emph{sequential} feature extractor in the front end and a supervised regressor in the back end. The sequential nature of the data is addressed by the recurrent network in the front, which acts as an automatic feature extractor for the following gradient-boosted supervised regressor. Therefore, a single hybrid architecture composed of a recurrent network and gradient boosted soft trees enjoying end-to-end optimization is proposed to effectively address the negative effects of separate feature selection and decision making processes for sequential data regression.

\subsection{Contributions}
We list our main contributions as follows.

\begin{enumerate}
    \item We introduce a hybrid architecture composed of an LSTM and a soft GBDT for sequential data prediction that can be trained in an end-to-end fashion with stochastic gradient descent. We also provide the backward pass equations that can be used in backpropagation.
    \item To the best of our knowledge, this model is the first in the literature that is armed with joint optimization for a sequential feature extractor in the front end and a supervised regressor in the back end.
    \item The soft GBDT regressor in the back end not only learns a feature mapping but also learns the optimal partitioning of the feature space.
    \item The proposed architecture is generic enough that the feature extractor part can readily be replaced with, for example, an RNN variant (e.g., gated recurrent unit (GRU) \cite{gru_paper}) or a temporal convolutional network (TCN) \cite{tcn_paper}. Similarly, the supervised regressor part can also be replaced with a machine learning algorithm as long as it is differentiable.
    \item Through an extensive set of experiments, we show the efficacy of the proposed model over real life datasets. We also empirically verify the integrity of our model with synthetic datasets.
    \item We publicly share our code for both model design and experiment reproducibility\footnote{https://github.com/mustafaaydn/lstm-sgbdt}.
\end{enumerate}

\subsection{Organization}
The rest of the paper is organized as follows. In Section \ref{sec:prelim}, we introduce the nonlinear prediction problem, describe the LSTM network in the front end and point to the unsuitableness of a \emph{hard} decision tree in joint optimization. We then describe the \emph{soft} GBDT regressor in the back end in Section \ref{sec:sGBDT} and the proposed end-to-end architecture along with the backward pass equations in Section \ref{sec:end2end}. We demonstrate the performance of the introduced architecture through a set of experiments in Section \ref{sec:Experiments}. Finally, Section \ref{sec:Conclusion} concludes the paper.

\section{Material and methods}\label{sec:prelim}

\subsection{Problem Statement}\label{sec:prob_state}
In this paper, all vectors are real column vectors and presented by boldface lowercase letters. Matrices are denoted by boldface uppercase letters. $x_k$ and $x_{t, k}$ denotes the $k\textsuperscript{th}$ element of the vectors $\boldsymbol{x}$ and $\boldsymbol{x}_t$, respectively. $\boldsymbol{x}^T$ represents the ordinary transpose of $\boldsymbol{x}$. $\boldsymbol{X}_{i, j}$ represents the entry at the $i\textsuperscript{th}$ row and the $j\textsuperscript{th}$ column of the matrix $\boldsymbol{X}$.

We study the nonlinear prediction of sequential data. We observe a sequence $\{y_t\}$ possibly along with a side information sequence $\{\boldsymbol{s}_t\}$. At each time $t$, given the past information $\{y_k\}$, $\{\boldsymbol{s}_k\}$ for $k \leq t$, we produce the predictions ${\boldsymbol{\hat d}}_H = [\,\hat y_{t+1}, \ldots, \hat y_{t+H}]^T$ where $H$ is the number of steps ahead to predict. Hence, in this purely online setting, our goal is to find the relationship
\begin{equation*}
    {\boldsymbol{\hat d}}_H = f\big([\,\ldots, y_{t-1}, y_t], [\,\ldots, \boldsymbol{s}_{t-1}, \boldsymbol{s}_{t}]\big),
\end{equation*}
where $f$ is a nonlinear function, which models ${\boldsymbol{ d}}_H = [\,y_{t+1}, \ldots, y_{t+H}]^T$. We note that $f$ can either produce $H$ estimates at once or do it in a recursive fashion (roll out) by using its own predictions. In either case, we suffer the loss over the prediction horizon given by
\begin{equation*}
    L = \frac{1}{H}\sum_{i=1}^{H} l(d_{H,i}, \hat{d}_{H,i}),
\end{equation*}
where $l$, for example, can be the squared error loss. As an example, in weather nowcasting, temperature ($y_t$) for the next six hours ($H=6$) are predicted using the hourly measurements of the past. Side information could include the wind and humidity levels ($\boldsymbol{s}_t \in \mathbbm{R}^2$). The model updates itself whenever new measurements of the next hour become available, i.e., works in an online manner.

We use RNNs for processing the sequential data. An RNN is described by the recursive equation \cite{rnn_itself}
\begin{equation*}
    \boldsymbol{h}_t = u(\boldsymbol{W}_{ih} \boldsymbol{x}_t + \boldsymbol{W}_{hh} \boldsymbol{h}_{t-1}),
\end{equation*}
where $\boldsymbol{x}_t$ is the input vector and $\boldsymbol{h}_t$  is the state vector at time instant $t$. $\boldsymbol{W}_{ih}$ and $\boldsymbol{W}_{hh}$ are the input-to-hidden and hidden-to-hidden weight matrices (augmented to include biases), respectively; $u$ is a nonlinear function that applies element wise. As an example, for the prediction task, one can use $\boldsymbol{s}_t$ directly as the input vector, i.e., $\boldsymbol{x}_t = \boldsymbol{s}_t$, or some combination of the past information $\{y_t, \ldots, y_{t-r+1}\}$ with a window size $r$ along with $\boldsymbol{s}_t$ to construct the input vector $\boldsymbol{x}_t$. The final output of the RNN is produced by passing $\boldsymbol{h}_t$ through a sigmoid or a linear model:
\begin{equation*}
    \boldsymbol{\hat d}_H = \boldsymbol{W} \boldsymbol{h}_t,
\end{equation*}
where $\boldsymbol{W}$ is $H \times m$ and $\boldsymbol{h}_t \in \mathbb{R}^m$.  Here, we consider $\boldsymbol{h}_t$ as the ``extracted feature vector'' from the sequential data by the RNN. 

To effectively deal with the temporal dependency in sequential data, we employ a specific kind of RNN, namely the LSTM neural networks \cite{lstm_itself} in the front end as a feature extractor. Here we use the variant with the forget gates \cite{lstm_forget}. The forward pass equations in one cell are described as
\begin{equation}\label{eq:lstmforward}
\begin{aligned}
    \boldsymbol{f}_t &= \sigma(\boldsymbol{W}_f \,\{\boldsymbol h_{t-1}, \boldsymbol x_t\} + \boldsymbol b_f)\\
    \boldsymbol{i}_t &= \sigma(\boldsymbol{W}_i \,\{\boldsymbol h_{t-1}, \boldsymbol x_t\} + \boldsymbol b_i)\\
    \tilde{\boldsymbol c}_t &= \tanh(\boldsymbol{W}_c \,\{\boldsymbol h_{t-1}, \boldsymbol x_t\} + \boldsymbol b_c)\\
    \boldsymbol{o}_t &= \sigma(\boldsymbol{W}_o \,\{\boldsymbol h_{t-1}, \boldsymbol x_t\} + \boldsymbol b_o)\\
    \boldsymbol{c}_t &= \boldsymbol{f}_t \odot \boldsymbol{c}_{t-1} + \boldsymbol{i}_t \odot \tilde{\boldsymbol c}_t\\
    \boldsymbol h_{t} &= \boldsymbol o_{t} \odot \tanh(\boldsymbol c_t),
\end{aligned}
\end{equation}
where $\boldsymbol c_t$ is the state vector, $\boldsymbol x_t$ is the input to the cell, $\boldsymbol h_{t-1}$ is the hidden state from the previous cell and $\{\boldsymbol h_{t-1}, \boldsymbol x_t\}$ denotes the vertical concatenation of $\boldsymbol x_t$ and $\boldsymbol h_{t-1}$. $\boldsymbol f_t, \boldsymbol i_t, \tilde{\boldsymbol c}_t, \boldsymbol o_t$ are the forget gate, input gate, block input and output gate vectors, respectively, whose stacked weight matrices are $\boldsymbol W_f, \boldsymbol W_i, \boldsymbol W_c$ and $\boldsymbol W_o$. The corresponding biases are represented by $\boldsymbol b_f, \boldsymbol b_i, \boldsymbol b_c$ and $\boldsymbol b_o$. $\sigma$ denotes the logistic sigmoid function, i.e, $\sigma(x) = 1 / (1 + \exp(-x))$ and $\tanh$ is the hyperbolic tangent function, i.e, $\tanh(x) = 2\sigma(2x) - 1$; both are nonlinear activation functions that apply point-wise. $\odot$ is the Hadamard (element-wise) product operator.
The constant error flow through the cell state in backpropagation allows LSTMs to prevent vanishing gradients, and nonlinear gated interactions described by \eqref{eq:lstmforward} control the information flow effectively to capture long term dependencies \cite{lstm_odssey}.

\begin{figure}[!t]
    \centering
    \begin{forest} 
    [D_1, tikz={
        \node at (0,0.3) (input) {$\boldsymbol{h}$};
        }
        [D_2, edge=blue
              [\phi_1] 
              [\phi_2, edge=blue,draw=blue,bottom color=green!50] 
        ]   
        [D_3
              [\phi_3] 
              [\phi_4] 
        ]   
    ] 
    \end{forest}
    \caption{A hard binary decision tree. Pink nodes are the internal nodes and green nodes are the leaf nodes. $D_i$'s represent the decision rules applied at each internal node to form a binary decision. For an example input $\boldsymbol{h}$, the set of decisions lead to the second leaf node where the path is colored blue. Therefore, $\boldsymbol{h}$ is associated with the value $\phi_2$ that is assigned to the second leaf node.}
    \label{fig:hardtree}
\end{figure}
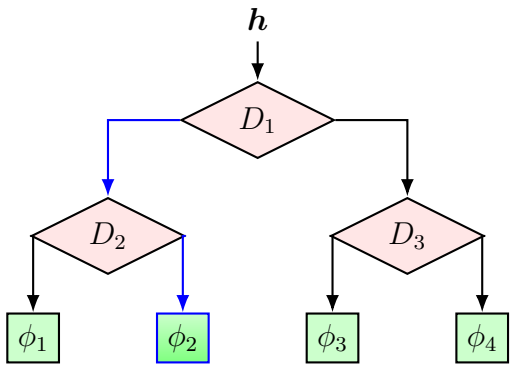

Instead of the linear model, one can use decision trees to process the extracted features, i.e., $\boldsymbol{h}_t$ produced by the RNN or LSTM.  A (hard) binary decision tree forms an axis-aligned partition of the input space \cite{hard_tree}. The tree consists of two types of nodes: internal nodes and leaf nodes, as depicted in Fig. \ref{fig:hardtree}. The input vector enters from the root node and is subjected to a decision rule at each internal node, routing it either to the left child or the right child. This path ends when the input ends up in a leaf node where an output value is produced, through which the input space is partitioned. Thereby, a regression tree with $L$ leaf nodes is recursively constructed to split the input space into $L$ regions, $R_\ell$, where $\ell = 1, 2, \ldots, L$. This construction is usually done with a greedy approach \cite{hard_tree}. The forward pass of an input vector $\boldsymbol{h}$ is given by
\begin{equation}\label{eq:hardtree}
    g(\boldsymbol{h}) = \sum_{\ell=1}^L \phi_\ell \mathbbm{1}(\boldsymbol{h} \in R_\ell),
\end{equation}
where $\phi_\ell$ is the scalar quantity that $\ell\textsuperscript{th}$ leaf node associates with the input, $\mathbbm{1}$ is the indicator function and $g$ represents the tree. By \eqref{eq:hardtree}, the summation has only one nonzero term due to the \emph{hard} decisions that route the input vector in a binary fashion, i.e., only one of the leaf nodes contributes to the final output.
A standalone tree, however, usually suffers from overfitting \cite{hard_tree} and a way to overcome this is to use many trees in a sequential manner, i.e., gradient boosting decision trees (GBDT) \cite{gbdt_friedman}. A GBDT works by employing individual trees as ``weak'' learners and fitting one after another to the residuals of the previous tree with the same greedy approach for constructing individual trees. The forward pass is the weighted sum of the predictions of each individual tree. For $M+1$ trees, the overall output for the input $\boldsymbol{h}$ is given by
\begin{equation}\label{eq:gbmgeneral}
    \text{GBM}(\boldsymbol{h}) = g^{(0)}(\boldsymbol{h}) + \sum_{j=1}^{M} \nu g^{(j)}(\boldsymbol{h}),
\end{equation}
where $g^{(j)}$ for $j = 0, \ldots, M$ represents the forward pass of the $j\textsuperscript{th}$ tree and $\nu$ is the shrinkage parameter that weights the predictions of individual trees (except for the first one in the chain) to provide regularization \cite{gbdt_friedman}. The very first tree, i.e., $g^{(0)}$, predicts a constant value regardless of the input, e.g., the mean value of the targets when the loss criterion is the mean squared error.

Connecting a recurrent neural network and a decision tree based architecture in an end-to-end fashion and training with gradient based methods is not possible when conventional decision trees are used as they inherently lack differentiability due to hard decisions at each node. Therefore, for the joint optimization, we introduce a model composed of an RNN in the front end and a \emph{soft} GBDT in the back end, which are jointly optimized in an end-to-end manner. We next introduce this architecture for nonlinear prediction.

\subsection{The proposed model}\label{sec:proposed}
We next introduce a model which jointly optimizes the feature selection and model building in a fully online manner. To this end, we use an LSTM network in the front end as a feature extractor from the sequential data and a soft GBDT (sGBDT) in the back end as a supervised regressor. Given $\{y_k\}$ and $\{\boldsymbol{s}_k\}$ for $k \leq t$ in the forward pass, a multi layer LSTM produces hidden state vectors by applying \eqref{eq:lstmforward} in each of its cells. Any differentiable pooling method can be used on the hidden state vectors of the last layer, e.g, last pooling where only the hidden state of the rightmost LSTM cell (when unrolled) is taken. This pooled quantity, $\boldsymbol{h}_\text{LSTM}$, is the filtered sequential raw data that represents the feature extraction or learning part via LSTM. This extracted feature vector $\boldsymbol{h}_\text{LSTM}$ is then fed into the sGBDT as the input to produce $\boldsymbol{\hat d}_H$. We now describe the soft decision tree based GBDT used in the model.

\subsubsection{sGBDT}\label{sec:sGBDT}
An sGBDT is a gradient boosting decision tree \cite{gbdt_friedman} where weak learners are soft decision trees \cite{irsoy_sdt,hinton_sdt}. In this section, we first study the soft decision tree variant we use in our model and then present the gradient boosting machinery that combines them.

\paragraph{Soft Decision Tree (sDT)}\label{sec:sdt}
The widely used \emph{hard} binary decision trees recursively route a sample from the root node to a single leaf node and result in (usually) axis-aligned partitions of the input space. The decision rules applied at each node redirect the sample either to the left or the right child. Soft decision trees differ in that they redirect a sample to \emph{all} leaf nodes of the tree with certain probabilities attached. These \emph{soft} decision rules yield a differentiable architecture.

\begin{figure}[!t]
    \centering
    \begin{adjustbox}{max width=0.91\linewidth}
    \begin{forest}
    before typesetting nodes={
          !r.replace by={[, coordinate, append]}
    },
    for tree={
        where n children=0{
          tier=word,bottom color=green!20,top color=green!20, inner sep=1.5mm,edge=blue,draw=blue,bottom color=green!50
        }{
          circle, aspect=2,bottom color=red!10,top color=red!10
        },
        edge+={thick, -Latex},
        math content,
        s sep'+=.7cm,
        draw,
        thick,
        edge path'={ (!u) -| (.parent)},
    }
    [{\boldsymbol{w}_1,b_1}, tikz={
        \node at (0,0.25) (input) {$\boldsymbol{h}$};
        }
        [{\boldsymbol{w}_2,b_2}, edge=blue
              [\boldsymbol\phi_1]
              [\boldsymbol\phi_2]
        ]   
        [{\boldsymbol{w}_3,b_3}, edge=blue
              [\boldsymbol\phi_3] 
              [\boldsymbol\phi_4]
        ]   
    ] 
    \end{forest}
    \end{adjustbox}
    \caption{A soft binary decision tree. Pink nodes are the internal nodes and green nodes are the leaf nodes, as in Fig. \ref{fig:hardtree}. Unlike the hard decision tree, however, the decision rule at each internal node is not hard and the input $\boldsymbol{h}$ is routed to both left \emph{and} right child, as indicated with blue edges. The direction is accompanied by a Bernoulli random variable parameterized by $\boldsymbol{w}_j$ and ${b}_j$. Therefore, \emph{all} of the leaf nodes has a contribution in the final prediction of the tree, as expressed in \eqref{eq:onesofttreeoutput}. We also note that the learnable parameters $\boldsymbol{\phi}_j$ in each leaf node are vector valued in general in contrast to hard decision trees.}
    \label{fig:softtree}
\end{figure}
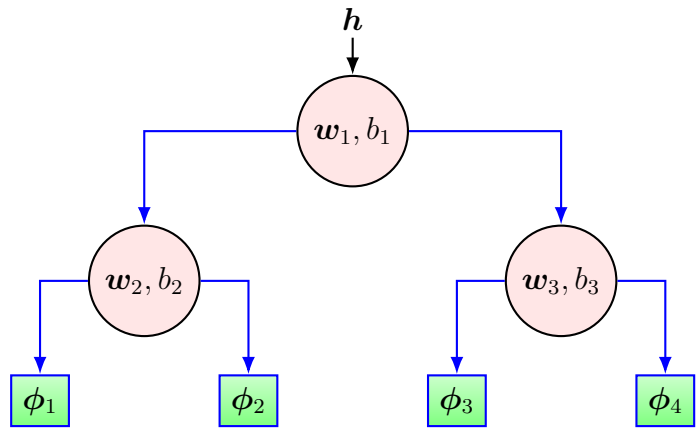

Formally, we have a binary tree with an ordered set of internal nodes $\mathcal{N}$ and leaf nodes $\mathcal{L}$. At the $m\textsuperscript{th}$ internal node $N_m \in \mathcal{N}$, the sample is subjected to a probabilistic routing controlled by a Bernoulli random variable with parameter $p_m$ where ``success'' corresponds to routing to the left child. A convenient way to obtain a probability is to use the sigmoid function $\sigma(z) = 1 / (1 + \exp(-z))$. Thereby we attach $\boldsymbol w_m$ and $b_m$ to each $N_m$ as learnable parameters such that $p_m = \sigma(\boldsymbol w_m^T \boldsymbol h + b_m)$ is the probability of sample $\boldsymbol h$ going to the left child and $1 - p_m$ is the probability of it going to the right child. With this scheme, we define a ``path probability'' for each leaf node $\ell \in \mathcal{L}$ as the multiplication of the probabilities of the internal nodes that led to it. As shown in Fig. \ref{fig:softtree}, \emph{every} leaf node has a probability attached to it for a given sample. These path probabilities correspond to each leaf nodes' contribution in the overall prediction given by the tree, which is
\begin{equation}\label{eq:onesofttreeoutput}
\begin{aligned}
    g(\boldsymbol{h}) &= \sum_{\ell \in \mathcal{L}} \text{pp}_\ell \,\boldsymbol \phi_\ell\\
    &= \sum_{\ell \in \mathcal{L}} \big(\hspace{-0.4em}\prod_{i \in \text{A}(\ell)} \hspace{-0.2em}p_i\big)\,\boldsymbol \phi_\ell\\
    &= \sum_{\ell \in \mathcal{L}}  \big(\hspace{-0.4em}\prod_{i \in \text{A}(\ell)} \hspace{-0.2em}\sigma(\boldsymbol w_i^T \boldsymbol h + b_i)\big)\,\boldsymbol \phi_\ell,
\end{aligned}
\end{equation}
where $\text{pp}_\ell$ is the path probability of leaf node $\ell$, $\boldsymbol \phi_\ell$ is the predicted value produced at $\ell$ which is a learnable parameter (and vector valued in general) and $\text{A}(\ell)$ denotes the ascendant nodes of the leaf node $\ell$, including the root node.

The soft decisions formed through $p_m$'s not only allow for a fully differentiable tree but also result in smooth decision boundaries that effectively reduce the number of nodes necessary to construct the tree in comparison to hard trees \cite{irsoy_sdt}. Furthermore, $\boldsymbol{\phi_\ell}$ being a vector valued quantity in general makes multi output regression possible.
\paragraph{Gradient boosting of sDTs}\label{sec:gbofsdts}
We adapt the gradient boosting machinery as originally proposed in \cite{gbdt_friedman}. The weak learners are fixed depth soft decision trees. Hence, all of the trees have a predetermined depth which saves computational time. In this regard, our sGBDT resembles AdaBoost where decision stumps are employed as weak learners \cite{adaboost_paper}. Note that our implementation can be extended to growing trees in a straightforward manner. As depicted in Fig. \ref{fig:tummodel}, each tree (except for the very first one) is fitted to the residuals of the previous tree in the boosting chain. The first tree, $g_0$, does not have any predecessor and is tasked to produce a constant output regardless of the input, which can be, for example, the mean of the target values of the training set. The forward pass equation of the sGBDT follows \eqref{eq:gbmgeneral}.  The first tree does not possess any learnable parameter and rather acts as a starting point in the boosting chain. The parameters of other trees, which include $\boldsymbol w_m, b_m$ for internal nodes and $\boldsymbol \phi_\ell$ for leaf nodes, are learned through backpropagation with gradient descent. The backward pass equations are given in the next section.

An sGBDT has three hyperparameters to tune: the number of trees, the depth of each tree and the shrinkage rate, which controls how much an individual tree contributes to the forwarded output. The number of trees and the shrinkage rate ideally move in an inversely proportional way, i.e., as the number of trees increases, one needs to reduce the shrinkage rate. The shrinkage rate not only acts as a learning rate parameter but also provides (inverse) regularization \cite{gbdt_friedman}. Therefore smaller values of shrinkage rate reduce overfitting; however, this leads to usage of large number of trees, increasing the computational cost. As the depth of each weak learner increases, not only the computational complexity increases exponentially but also the risk of overfitting arises. On the other hand, stumps might easily underfit a given dataset. To remedy these trade-offs, we employ validation procedures to choose these hyperparameters for a given task. As an example, we perform a k-fold cross validation over the training split of the data and determine the ``best'' configuration of hyperparameters.
\subsubsection{The end-to-end model}\label{sec:end2end}
\begin{figure*}[!t]
\centering
\includegraphics[width=\textwidth]{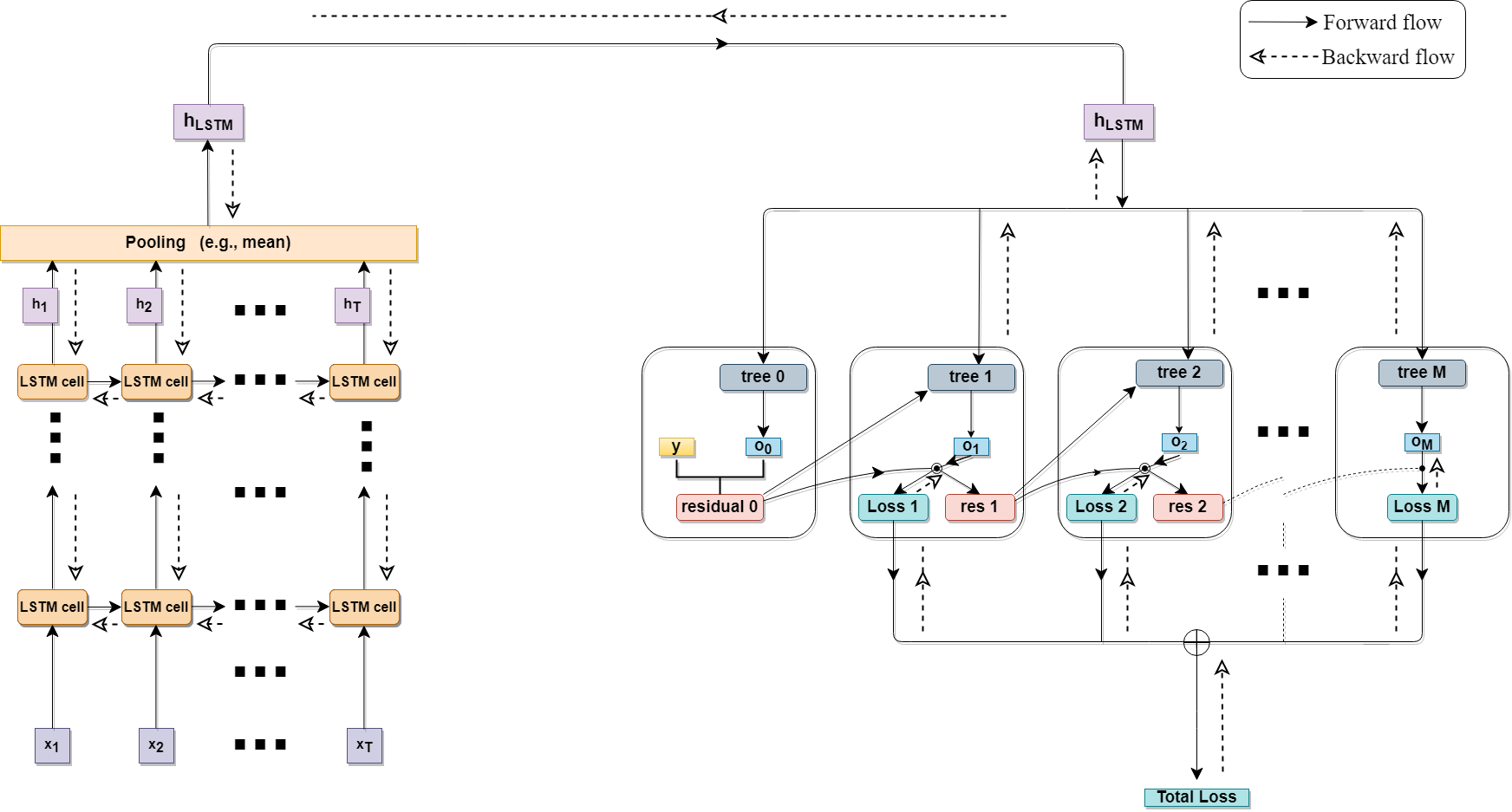}
\caption{The proposed LSTM-SGBDT architecture. The recurrent network in the front end is tasked to extract a feature vector, $\boldsymbol h_{\text{LSTM}}$, from the raw sequential data. This extracted feature vector is a summary (i.e., pooling) of the last layer's hidden states of all time steps. $\boldsymbol h_{\text{LSTM}}$ is then fed to the soft GBDT in the back end where the weak learners are soft decision trees. Boosted trees act as a supervised regressor; however, the need for hand-designed features are mitigated. The total loss is backpropagated through the whole architecture, whose direction is shown by the dashed lines, achieving an end-to-end optimization.}
\label{fig:tummodel}
\end{figure*}

Here, we describe how a given sequential data is fed forward and the corresponding backward pass equations of the joint optimization. Given an $L$-layer LSTM network, the input is first windowed and then fed into the first layer of LSTM, denoted by $\text{LSTM}\textsuperscript{(1)}$. The window size $T$ determines how many time steps the model should look back for the prediction, i.e., the number of steps an LSTM cell should be unrolled in each layer. Each cell applies \eqref{eq:lstmforward} sequentially where $\boldsymbol h_0 = \boldsymbol c_0 = \boldsymbol 0$, i.e., initial cell gets $\boldsymbol 0$ vector for both the hidden state and cell state. Hidden states at all time steps ($\boldsymbol h_t^{(j)}$ for the $j\textsuperscript{th}$ layer) are recorded and fed into the next layer of LSTM as the inputs as shown in the left side of Fig. \ref{fig:tummodel}. This process repeats until it reaches the last layer of LSTM whereby a pooling method, $P$, is applied to the hidden states of each cell $\boldsymbol h_t^{(L)} \text{ for } t = 1, \ldots, T$. $P$ can be any differentiable operation; here we present three options of \emph{last}, \emph{mean} and \emph{max} pooling as
\begin{equation}\label{eq:pools}
\begin{aligned}
    P_{last}(\boldsymbol{h}_1^{(L)}, \ldots, \boldsymbol{h}_{T}^{(L)}) &= \boldsymbol{h}_{T}^{(L)}\\
    P_{mean}(\boldsymbol{h}_1^{(L)}, \ldots, \boldsymbol{h}_{T}^{(L)}) &= \frac{1}{T}\sum_{t=1}^{T}\boldsymbol{h}_t^{(L)}\\
    P_{max}(\boldsymbol{h}_1^{(L)}, \ldots, \boldsymbol{h}_{T}^{(L)})_i &= \max_{t} h_{t, i}^{(L)}.
\end{aligned}
\end{equation}
The pooled LSTM output, $\boldsymbol h_{\text{LSTM}} = P(\boldsymbol{h}_1^{(L)}, \ldots, \boldsymbol{h}_{T}^{(L)})$, carries the features extracted from the raw sequential data and becomes the input vector for the sGBDT, as shown in Fig. \ref{fig:tummodel}. Then, as per the gradient boosting machinery, $\boldsymbol h_{\text{LSTM}}$ is fed into $M + 1$ soft trees, last $M$ of which give an output as described in \eqref{eq:onesofttreeoutput}, whereas the first one outputs a constant value irrespective of the input, as explained in Section \ref{sec:gbofsdts}. Nevertheless, we still call this first component in the chain a \emph{tree}. The output of the sGBDT is a weighted sum of individual tree outputs. Combining \eqref{eq:gbmgeneral} and \eqref{eq:onesofttreeoutput}, the overall output of the sGBDT is given by
\begin{equation}\label{eq:sgbdtoutput}
\begin{aligned}
    \text{sGBDT}(\boldsymbol{h}) &= \sum_{j=0}^{M}\nu^{(j)} g^{(j)}(\boldsymbol{h})\\
    &= g^{(0)}(\boldsymbol{h}) + \nu \sum_{j=1}^{M} \sum_{\ell \in \mathcal{L}^{(j)}} \prod_{i \in \text{A}(\ell^{(j)})} p_i^{(j)} \boldsymbol \phi_\ell^{(j)},
\end{aligned}
\end{equation}
where
\begin{equation*}
    \nu^{(j)} = 
    \begin{cases}
    1, & \text{if } j = 0 \\
    \nu, & \text{otherwise},
    \end{cases}
\end{equation*}
$g^{(j)}$ represents the forward pass of the $j\textsuperscript{th}$ tree, $\mathcal{L}^{(j)}$ is the set of leaf nodes in the $j\textsuperscript{th}$ tree and $\text{A}(\ell^{(j)})$ is the set of ascendant nodes of leaf node $\ell$ of the $j\textsuperscript{th}$ tree. We note that the superscript $(j)$ indicates that the quantity belongs to the $j\textsuperscript{th}$ tree. This completes the forward pass of the data.

We fit this architecture in an end-to-end manner using backpropagation with gradient descent. For the backward pass equations, without loss of generality, we focus on the case where values produced at leaf nodes of the trees, $\phi_\ell^{(j)}$, are scalar quantities; and we use the mean squared error for the loss function. Further, for brevity, we assume a 1-layer LSTM and let $\boldsymbol{\tilde{h}}_{\text{LSTM}}$ be the augmented version of the pooled LSTM output $\boldsymbol h_{\text{LSTM}}$ (i.e., it has a $1$ prepended to it so that $\boldsymbol w_m$ and $b_m$ described in Section \ref{sec:sdt} are collapsed into one vector, $\boldsymbol w_m$). Let $\boldsymbol{x}_{t'} \in \mathbb{R}^{m+1}$ with $t' = 1, \ldots, T$ be the input vector (possibly augmented with side information $\boldsymbol{s}_{t'} \in \mathbb{R}^m$; e.g., $\boldsymbol{x}_{t'} = [y_{t+1-t'}, \boldsymbol{s}_{t'}^T]^T)$ fed into $t'\textsuperscript{th}$ LSTM cell when unrolled and $y_{true} \coloneqq y_{t+1}$ be the ground truth, which is a real number. The total loss $E$ obtained from the soft trees, as shown in Fig. \ref{fig:tummodel}, can be written as
\begin{equation*}
    E = \text {Total Loss} = \sum_{j=1}^M \text{Loss}^{(j)} = \sum_{j=1}^M (r^{(j)} - \nu o^{(j)})^2,
\end{equation*}
where $o^{(j)}$ is the output of the $j\textsuperscript{th}$ tree as in \eqref{eq:onesofttreeoutput}, $\nu$ is the shrinkage rate parameter that helps for regularization \cite{gbdt_friedman} and $r^{(j)}$ is the residual that becomes the supervised signal the $j\textsuperscript{th}$ tree is fitted to, i.e., 
\begin{align*}
    o^{(j)} &= \sum_{\ell^{(j)}} \text{pp}_\ell ^ {(j)} \phi_{\ell}^{(j)}\\
    r^{(j)} &= y_{{true}} - \nu \sum_{i=0}^{j-1} o^{(i)}.
\end{align*}
We now present the gradient of the total loss with respect to the learnable parameters. For the sGBDT, we have $\phi_\ell^{(j)}$ for each leaf node $\ell^{(j)}$ and $\boldsymbol{w_m}^{(j)}$ for each internal node $m^{(j)}$ in the $j\textsuperscript{th}$ tree. The gradients are given as
\begin{align}
\frac{\partial E}{\partial \phi_{\ell}^{(j)}} &= \text{pp}_\ell ^ {(j)} \sum_{k=j}^M 2(\nu o^{(k)} - r^{(k)})\label{eq:gradwrtsoftparamsphi}\\
\begin{split}\label{eq:gradwrtsoftparamsw}
\frac{\partial E}{\partial {\mathbf w}_{m}^{(j)}} &= \bigg(\sum_{\ell^{(j)} \in \text D(\text m^{(j)})} \phi_{\ell}^{(j)}\frac{\text{pp}_\ell ^ {(j)} \text{p}_m ^ {(j)} (1-\text{p}_m ^ {(j)})}{\text{p}_m ^ {(j)} -  \mathbbm{1}(\ell^{(j)} \in \text {LD}( m^{(j)}))}\\
&\qquad\qquad\qquad \boldsymbol{\tilde{h}}_{\text{LSTM}} \; \sum_{k=j}^M 2(\nu o^{(k)} - r^{(k)}) \bigg),
\end{split}
\end{align}
where $M$ is the number of weak learners (soft decision trees), $\text D(m)$ is the set of descendants of the internal node $m$, $\text {LD}(m)$ is the set of \emph{left} descendants of the internal node $m$, i.e., it consists of the nodes in the sub-tree rooted at node $m$ which are reached from $m$ by first selecting the left branch of $m$. \eqref{eq:gradwrtsoftparamsphi} and \eqref{eq:gradwrtsoftparamsw} give the necessary components for the update equations of the learnable parameters of the sGBDT, namely, $\boldsymbol{w}_m^{(j)}$ and $\phi_\ell^{(j)}$ for each tree and node.

For the joint optimization, we also derive the gradient of the loss with respect to $\boldsymbol{\tilde{h}}_{\text{LSTM}}$, which is given as
\begin{equation}\label{eq:gradwrthLSTM}
    \begin{split}
        &\frac{\partial E}{\partial \boldsymbol{\tilde{h}}_{\text{LSTM}}} =\\
        &\qquad\sum_{j=1}^M \Bigg(\sum_{\ell^{(j)}} \phi_{\ell}^{(j)} \text{pp}_\ell ^ {(j)}\\
        &\qquad\qquad\biggl(\sum_{n^{(j)} \in \text A(\ell ^{(j)})} (\mathbbm{1}(n^{(j)} \in \text {RA}(\ell^{(j)})) - p_n^{(j)}) {\mathbf w}_n^{(j)} \bigg)\\
        &\qquad \qquad \qquad\sum_{k=j}^M 2(\nu o^{(k)} - r^{(k)})\Bigg),
    \end{split}
\end{equation}
where $\text A(\ell)$ is the set of ascendants of the leaf node $\ell$, $\text {RA}(\ell)$ is the set of \emph{right} ascendants of the leaf node $\ell$, i.e., it consists of the nodes in the tree from which by first selecting the right branch, there is a path that leads to $\ell$. With \eqref{eq:gradwrthLSTM}, the backpropagation path until the pooled output of LSTM is completed.

We assume last pooling, i.e., $P = P_{last}$ of \eqref{eq:pools} for the derivation such that the error propagates through the hidden state of the last cell. The gradients of the total loss with respect to LSTM parameters are given by
\begin{align}
    \frac{\partial E}{\partial \boldsymbol{W}_\star} &= \sum_{t = 1}^{T}\big(\frac{\partial E}{\partial \boldsymbol{h}_t} \odot \delta\boldsymbol{\!\star}_t\big)\,\tilde{\boldsymbol{x}}_t^T\label{eq:lstmgradweight}\\
    \frac{\partial E}{\partial \boldsymbol{b}_\star} &= \sum_{t = 1}^{T} \frac{\partial E}{\partial \boldsymbol{h}_t} \odot \delta\boldsymbol{\star}_t\label{eq:lstmgradbias},
\end{align}
where $\star$ denotes the set $\{\hspace{-0.1em}f, i, c, o\}$,
\begin{equation*}\label{eq:lstmgradhiddenstate}
    {\frac{\partial E}{\partial \boldsymbol{h}_{t-1}^T}} = \frac{\partial E}{\partial \boldsymbol{h}_{t}} (\Lambda(\delta\boldsymbol{o}_t)\tilde{\boldsymbol{W}}_o + \Lambda(\delta\hspace{-0.1em}\boldsymbol{f}_t)\tilde{\boldsymbol{W}}_f     +\Lambda(\delta\boldsymbol{i}_t)\tilde{\boldsymbol{W}}_i +  \Lambda(\delta\boldsymbol{c}_t)\tilde{\boldsymbol{W}}_c)^T,
\end{equation*}
starting from the hidden state of the rightmost cell (i.e, $t = T$) with ${\partial E}/{\partial \boldsymbol{h}_{T}} = {\partial E}/{\partial \boldsymbol{\tilde{h}}_{\text{LSTM}}}$ given by \eqref{eq:gradwrthLSTM} and $\tilde{\boldsymbol{W}}_\star$ denotes the square matrix inside the stacked ${\boldsymbol{W}}_\star$ that multiplies $\boldsymbol{h}_{t-1}$ in \eqref{eq:lstmforward}. $\Lambda(\cdot)$ results in a square, diagonal matrix where the diagonal entries are given by its vector operand. Finally, $\delta\,\boldsymbol{\!\star}_t$ are given by
\begin{equation*}\label{eq:lstmgradhelpers}
\begin{aligned}
    \delta\hspace{-0.1em}\boldsymbol{f}_t &= \boldsymbol{o}_t \odot \tanh'({\boldsymbol{c}_t}) \odot \boldsymbol{c}_{t-1}\odot\sigma'(\hspace{-0.1em}\boldsymbol{f}_t)\\
    \delta\boldsymbol{i}_t &= \boldsymbol{o}_t \odot \tanh'({\boldsymbol{c}_t}) \odot \tilde{\boldsymbol{c}}_{t}\odot\sigma'(\boldsymbol{i}_t)\\
    \delta\boldsymbol{c}_t &= \boldsymbol{o}_t \odot \tanh'({\boldsymbol{c}_t}) \odot \boldsymbol{i}_{t}\odot(1 - \tilde{\boldsymbol{c}}_{t}^2)\\
    \delta\boldsymbol{o}_t &= \tanh({\boldsymbol{c}_t}) \odot\sigma'(\boldsymbol{o}_t).
\end{aligned}
\end{equation*}

\Crefrange{eq:gradwrtsoftparamsphi}{eq:lstmgradbias} give the required backward pass equations of the whole architecture. The corresponding stochastic gradient update equations are
\begin{align}\label{eq:sgdupdates}
    \Delta \phi_{\ell}^{(j)} &= -\eta \,\frac{\partial E}{\partial \phi_{\ell}^{(j)}} &
    \Delta \boldsymbol w_m^{(j)} &= -\eta \,\frac{\partial E}{\partial \boldsymbol w_m^{(j)}}\\
    \Delta \boldsymbol{W}_\star &= -\eta \,\frac{\partial E}{\partial \boldsymbol{W}_\star} &
    \Delta \boldsymbol{b}_\star &= -\eta \,\frac{\partial E}{\partial \boldsymbol{b}_\star},
\end{align}
where $\Delta$ represents the change in the variable between two successive iterations and $\eta$ is the learning rate hyperparameter.

\textit{\textbf{Remark.}} Given a conventional LSTM network described by \eqref{eq:lstmforward}, for an input vector of dimension $m$ and hidden state dimension $q$, there are four matrix-vector multiplications for the input, four matrix-vector multiplications for the recurrent state, and three vector-vector multiplications between the gates; these correspond to $4q^2 + 4qm + 3q$ multiplication operations in total. On the other hand, a soft tree of depth $d$, which is fed with the resultant state vector of the LSTM in the front end (of dimension $q$) has $2^d - $ internal nodes and therefore results in $q(2^d - 1)$ multiplications for the probability calculations and $2^d - 1$ multiplications for the path probabilities, leading to $(q + 1)(2^d - 1)$ multiplications in total per tree. Hence, the total number of multiplications an sGBDT with $M$ trees brings is $M (2^d - 1)(q + 1)$. Although the depth seems to increase complexity exponentially, since it is constant per tree and does not exceed 3 in applications, it can be treated as a constant along with $M$. Overall, the multiplicational complexity of the hybrid model remains in the order of $q^2$ in terms of the hidden state dimension of the LSTM network, which is the same as using an LSTM alone.

\section{Experiments}\label{sec:Experiments}
In this section, we illustrate the performance of the proposed architecture under different scenarios. In the first part, we demonstrate the learning advantages of the proposed model with respect to disjoint structures through synthetic datasets. We also perform an ablation study to exhibit the importances of each component of the model. We then consider the regression performance of the model over well-known real life datasets such as bank \cite{bank_dataset}, elevators \cite{elev_puma_dataset}, kinematics \cite{kine_dataset} and pumadyn \cite{elev_puma_dataset} datasets in an online setting. We then perform simulations over the hourly, daily and yearly subsets of the M4 competition \cite{m4_comp} dataset, aiming one-step ahead forecasting. We lastly consider two financial datasets, Alcoa stock price \cite{alcoa_dataset} and Hong Kong exchange rate data against USD \cite{hk_dataset} where we replace the LSTM part of our model with a GRU network to show the generic nature of our framework. In each experiment, for all models, the only preprocessing step we applied to the datasets is standardization. In offline settings (e.g., with the M4 dataset), we first extract the mean and the standard deviation of the features across the training split of the data, and scale both the training and testing data with those same statistics, i.e., for each feature, we subtract the mean and divide by the standard deviation. In online settings (e.g., in financial datasets), we apply standardization cumulatively, i.e., we store a running mean and running standard deviation obtained from the available data for scaling and update them as new data points arise. We use standardization to prevent possible scaling differences among the raw features, to which the gradient descent is highly sensitive and it could lead to performance degradation in learning \cite{scaling_backprop}. We emphasize that we do not extract any additional features from the datasets; the LSTM network in the front end is tasked to automatically extract the features from the raw data. We use the past values of the target signal to feed into the LSTM network, and use the side information features whenever dataset provides them, e.g., in the real life datasets of Section 3.2. All the experiments were conducted using a computer with i7-6700HQ processor, 2.6 GHz CPU and 12 GB RAM. 

\subsection{Architecture Verification}\label{sec:verify_integ}
\begin{figure}[!t]
\centering
\includegraphics[width=0.9\linewidth]{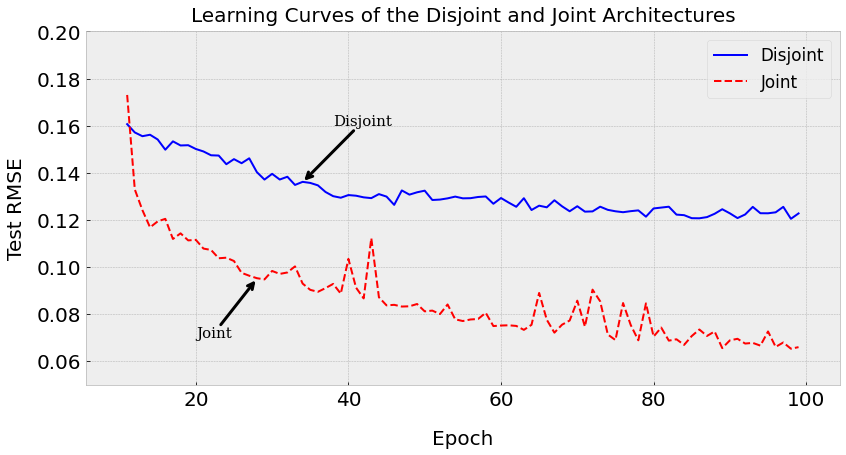}
\caption{Learning curve comparison with a disjoint architecture. The joint architecture not only mitigates training separate models but also achieves closer to optimal performance compared to the disjoint architecture.}
\label{fig:verify_arch_2}
\end{figure}

In this section, we show the advantages of our end-to-end optimization framework with respect to the classical disjoint framework, i.e., where one first extracts features and then optimizes the regressor. For controlled experiments, we first work with artificially generated data that are generated through an LSTM network. The inputs are from i.i.d. standard Gaussian distribution and the targets, $y_t$, are formed as the output of a fixed LSTM network with a given seed. The ``disjoint'' model consists of an LSTM for feature extraction and a hard GBDT for regressing over those features. We note that the mechanism, as stated in \cite{disj_paper}, requires fitting the models independently. On the other hand, our ``joint'' model has an LSTM in the front end and a soft GBDT in the back end, while the optimization is now end-to-end. We generate 1,000 artificial samples with a fixed seed and spare 20\% of the data to test set. Both models are trained for 100 epochs. At the end of each epoch, we subject both models to test data and record their prediction performance in RMSE. The results are seen in Fig. \ref{fig:verify_arch_2}. We observe that the disjoint model struggles to reach an optimal solution and stabilizes its performance after around 40 epochs while the joint architecture has steadily-decreasing error performance on the test set. This shows the advantage of employing end-to-end optimization and verifies the working characteristics of our algorithm.

We further test the integrity of our end-to-end model using datasets that are produced using both LSTM and GBDT. Through this, we check the learning mechanism of the model in various controlled scenarios. First, we generate a random sequential input from an i.i.d. standard Gaussian distribution and feed it o an LSTM network whose weights are fixed with a random seed. Its output is then fed to a fixed soft GBDT to get the ground truths, i.e, $y_t$. The architecture is then subjected to \emph{replicate} these ground truth values as close as possible given the same random inputs. This tests whether the model is able to tune its parameters in an end-to-end manner to fit the given synthetic dataset. Secondly, we feed the above randomly generated data directly into a soft GBDT and get the ground truths as its output. We then train the model with this input-output pair, expecting that the LSTM part in the front end learns the \emph{identity} mapping. Lastly, we again use a soft GBDT to generate the same input-output pair, but train the overall model with the inputs $\boldsymbol{A}\boldsymbol{X}$ where $\boldsymbol{X}$ is the randomly generated design matrix and $\boldsymbol{A}$ is a random constant matrix of suitable shape. This configuration aims for LSTM to learn an \emph{inverse} mapping. We generate 1,000 samples in all three configurations.

\begin{figure}[!t]
\centering
\includegraphics[width=0.9\linewidth]{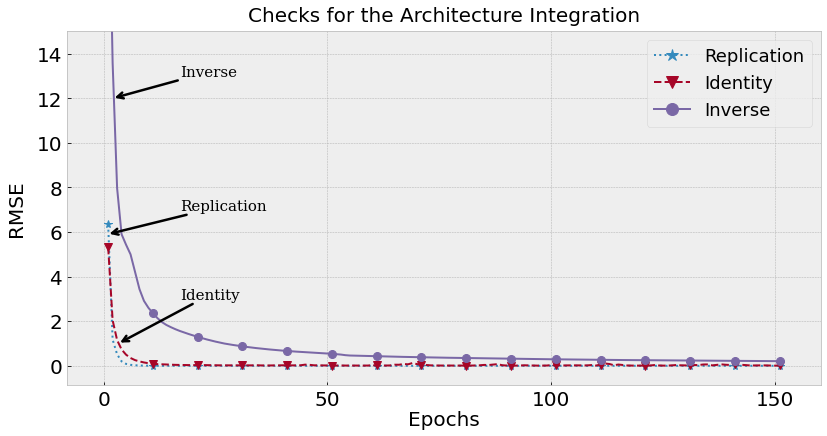}
\caption{Integrity verification results of the hybrid model for different configurations.}
\label{fig:verify_arch_1}
\end{figure}

In Fig. \ref{fig:verify_arch_1}, we observe that in all configurations, the network reaches to near zero root mean squared error (RMSE) between its predictions and the ground truths. Convergence is very fast in \emph{replication} and \emph{identity} tasks suggesting that the model is capable of learning from the individual parts and that the joint optimization readily tunes the parameters of the LSTM network. In the \emph{inverse} task, the convergence is rather slow since the input was subjected to a random projection. Still, the LSTM in the front end learns an inverse mapping, verifying the integrity of the end-to-end model.

\subsubsection{Ablation Study}\label{sec:ablation}
In order to show the importance of both feature extractor and supervised regressor parts of the proposed model, we perform an ablation study. To this end, we train three model configurations. First one is the fully functional LSTM-SGBDT model, where the feed-forward and backward update operations are as described in Section \ref{sec:proposed}. Second model is an LSTM-SGBDT where the feed-forward mechanism remains intact but the feature extractor LSTM part is ``frozen'', i.e., the corresponding learned parameters are \emph{not} updated through backpropagation. The parameters of the SGBDT part, however, are still updated through backpropagation. Similarly, the third model freezes its supervised regressor SGBDT part. We use 255 randomly chosen hourly time series from the M4 competition dataset \cite{m4_comp} and perform one-step ahead forecasting with all three models. We report the mean absolute percentage error (MAPE) over the 48 hours of forecast horizon as it is scale-independent and defined as
\begin{equation*}
    \text{MAPE}({\boldsymbol{d}}_H, {\boldsymbol{\hat d}}_H) = \frac{1}{H}\sum_{i=1}^{H}\, \abs{\frac{d_{H,i} - \hat{d}_{H,i}}{d_{H,i}}},
\end{equation*}
where $H$ is the prediction horizon (e.g., 48 hours), ${\boldsymbol{d}}_H$ and ${\boldsymbol{\hat d}}_H$ denote the vector of true and predicted values over the horizon, respectively. For a fair comparison, we use the same number of epochs in each model's training phase. Other hyperparameters are validated with 3-fold cross validation over the training split of each series. We also report the timing of each model; we note that the duration of validation in timings is not included, i.e., only the elapsed times over training and prediction of the ``best'' model is reported for each three models. In Table \ref{table:ablations}, we observe that the full model is more than three times better than either of the frozen models in mean MAPE performance over 255 time series, which shows that both the feature extractor LSTM part and the supervised regressor part SGBDT are crucial parts of the proposed model. In addition, the timing of the fully functional model is, as expected, more than the frozen models; however, the difference is tolerable as it is around 3 to 6 seconds over 255 series, making it a favorable choice.
\begin{table}[!t]
    \centering
    \begin{tabular}{c c c} 
         & Mean MAPE & Mean Time (s)  \\
        \hline
        Model I (full)           & \textbf{0.1837} & 15.42  \\ 
        Model II (frozen LSTM)   & 0.6120 & \textbf{9.83} \\
        Model III (frozen SGBDT) & 0.5696 & 12.33 \\
        \hline
    \end{tabular}
    \caption{Mean MAPE metrics and timings of three models for the ablation study over 255 hourly time series from the M4 dataset.}
    \label{table:ablations}
\end{table}
\subsection{Real Life Datasets}\label{sec:real_life_datasets}
In this section, we evaluate the performance of the proposed model in an online learning setting. To this end, we consider four real life datasets.

\begin{itemize}
	\item Bank dataset \cite{bank_dataset} is a simulation of queues in a series of banks using customers with varying level of complexity of tasks and patience. We aim to predict the fraction of customers that have left the queue without getting their task done. There are 8,192 samples and each sample has 32 features, i.e., $\boldsymbol{x}_t \in \mathbb{R}^{32}$.
	\item Elevators dataset \cite{elev_puma_dataset} is obtained from the experiments of controlling an F16 aircraft; the aim is to predict a numeric variable in range $[0.012, 0.078]$ related to an action on the elevator of the plane, given 18 features. The dataset has 16,599 samples.
	\item Kinematics dataset \cite{kine_dataset} consists of 8,192 samples that are a result of a realistic simulation of the dynamics of an 8-link all-revolute robot arm. We aim to predict the distance of the end-effector from a given target using 8 features.
	\item Pumadyn dataset \cite{elev_puma_dataset}, similar to Kinematics dataset, contains a realistic simulation of a Puma 560 robot arm. The goal is to predict the angular acceleration of one of the robot arm's links. There are 8,192 samples and 32 features.
\end{itemize}

We compare our model against six different models: the Naive model, autoregressive integrated moving average model with exogenous regressors (ARIMAX), multilayer perceptron (MLP), light gradient boosting machine (LightGBM), the conventional LSTM architecture and the disjoint model composed of an LSTM and a hard tree-based gradient boosting machine as described in Section \ref{sec:verify_integ}. The Naive model predicts $\hat y_t = y_{t-1}$ for all time steps, i.e., it carries the last measurement as its prediction for the next time step. ARIMAX model, as mentioned in Section \ref{sec:backg}, is a statistical model fitting an infinite impulse response (IIR) filter to the (possibly differenced) data with side information. We tune the number of AR and MA components using a stepwise algorithm outlined in \cite{auto_arima}. The MLP we use in our experiments approximate a nonlinear autoregressive process with exogenous regressors (ARX), i.e., the values of the input neurons are the lagged values of the target signal as well as any side information the data provides. We use a one-layer shallow network for which we tune number of hidden neurons, learning rate and the activation function in the hidden layer (ReLU \cite{relu} is used for the output neuron). LightGBM is, as mentioned in Section \ref{sec:backg}, a hard tree-based gradient boosting machine, which, with well-designed features, achieves effective sequential data prediction. We tune the number of trees, learning rate, bagging fraction, maximum number of leaves and regularization constants for LightGBM. For the LSTM model, we tune the number of layers, number of hidden units in each layer and the learning rate. For our architecture, we also search for the pooling method $P$, number of soft trees $M+1$, depth of each tree and the shrinkage rate $\nu$ in boosting. Among these six models, however, only the Naive model, MLP and LSTM are amenable to online learning; for the other 3 models we proceed as follows. For ARIMAX, we use the underlying Kalman filter to perform training; for LightGBM and the disjoint model, we would ideally perform re-training over the accumulated data at each time step; however, since this would already become time consuming, we introduce a queue of fixed size, say 50, and when it is fully filled, we append the new data to the historical data to refit the model; the queue is then emptied.

To assess the performance of online regression, we measure time accumulated mean squared errors of models, which is computed as the cumulative sum of MSE's normalized by the data length at a given time $t$. In Table \ref{table:real_life}, we report the cumulative error at the last time step for all models. We do not apply any preprocessing to the data for the Naive, ARIMAX and LightGBM models, while applying the standardization described in Section \ref{sec:Experiments} to other four models.
\begin{table}[!b]
    \centering
    \begin{tabular}{c c c c c} 
                   & Bank       & Elevators      & Kinematics      & Pumadyn  \\
        \hline
        Naive      & 0.0307     &  0.0714        &  0.1305         &  0.0218  \\   
        ARIMAX     & 0.0183     &  0.0465        &  0.0897         &  0.0107  \\
        MLP        & 0.0165     &  0.0043        & \textbf{0.0766} &  0.0019  \\
        LightGBM   & 0.0203     &  0.0234        &  0.0991         &  0.0143  \\
        LSTM       & 0.0190     &  0.0119        &  0.0884         &  0.0094  \\
        Disjoint   & 0.0236     &  0.0650        &  0.1102         &  0.0160  \\
        LSTM-SGBDT & \textbf{0.0151} & $\mathbf{2.671\times10^{-5}}$ & 0.0787 & \textbf{0.0009} \\
        \hline
    \end{tabular}
    \caption{Cumulative errors of the models on Bank, Elevators, Kinematics and Pumadyn datasets.}
    \label{table:real_life}
\end{table}

We observe that our joint model performs better on almost all datasets (except for Kinematics) over the last cumulative error metric against other models. In particular, in the Elevators dataset where the sample size is the largest, the difference in normalized cumulative error compared to other models is the greatest. This suggests that the baseline models are unable to learn from new data in long term as well as the joint architecture, featuring the role of the embedded soft GBDT as the final regressor. Similarly, in the Pumadyn dataset where the number of features is the largest, there is an order of magnitude difference in performances (except for MLP; the performance is still as twice as better). This verifies the importance of mitigating hand-designed features and letting LSTM in the front end to extract necessary information from raw features to feed to the regressor in the end. In addition, our end-to-end model consistently achieves much better performance than the disjoint architecture over all datasets, featuring the importance of joint optimization. In all datasets, Naive model has the highest errors as expected, and LightGBM model performs worse than MLP on all datasets; this is mainly because LightGBM is not tailored for online learning and required fully refitting, while MLP is amenable to do online updates using, for example, SGD. MLP performs slightly better than our model in Kinematics dataset; since MLP we used approximates a nonlinear autoregressive process, this implies that there might be high partial autocorrelation patterns in the particular dataset which MLP specifically focused on. Overall, the hybrid model achieves better than both its disjoint counterpart and various standalone models including LSTM over real life datasets, exhibiting its effectiveness in an online setting.

\subsection{M4 Competition Datasets}\label{sec:m4_datasets}
The M4 competition \cite{m4_comp} provided 100,000 time series of varying length with yearly, quarterly, monthly, weekly, daily and hourly frequencies. In our experiments, we chose yearly, daily and hourly datasets. The yearly dataset consists of 23,000 very short series with the average length being 31.32 years. Therefore, inclusion of this dataset aims to assess the performance of the models under sparse data conditions. Next, the daily dataset has 4,227 very long series where the average length is 2357.38 days. This dataset helps assess how effective a model is in capturing long-term trends and time shifts. Lastly, the hourly dataset consists of 414 series. The reason for including this subset is the dominant seasonal component. We experiment over all of the hourly series and randomly selected 500 of daily and yearly series each. The prediction horizon for hourly, daily and yearly datasets are 48, 14 and 6 time steps, respectively. We aim for one-step ahead predictions in an offline setting; we train and validate the models over the training split of the data, and then predict the test split of the data one step at a time \emph{without} refitting/updating the models with the arriving of new data.

We compare our model (LSTM-SGBDT) with six models presented in Section \ref{sec:real_life_datasets}; except we use seasonal ARIMAX (SARIMAX) instead of ARIMAX in the hourly dataset to account for the seasonality (we report the model's name as SARIMAX in Table \ref{table:m4} for brevity, though no seasonal component was fitted for yearly and daily datasets). For each dataset, we separate a hold-out subseries from the end of each time series as long as the desired prediction horizon. For example, a series in the hourly dataset has its last 48 samples spared for validation of the hyperparameters. The searched hyperparameters of models are as described in Section \ref{sec:real_life_datasets}. We use MAPE to evaluate models' performance across series. The preprocessing setup is the same as in the experiments of Section \ref{sec:real_life_datasets}.

The mean MAPE scores of the models across each datasets are seen in Table \ref{table:m4}. The LSTM-SGBDT model achieves better performance than other models in hourly and yearly datasets; in the daily dataset, LightGBM has a slightly better error metric. In the yearly dataset, which has short-length series, LSTM-SGBDT model struggles less than other models against the sparse data. This indicates that soft GBDT part helps reduce overfitting of the data hungry LSTM network and acts as a regularizer. LSTM-SGBDT also outperforms the vanilla LSTM in the daily dataset with a wider margin of $74.3\%$. Given that the daily dataset consists of very long series, LSTM-SGBDT is able to model long-term dependencies better than LSTM. Lastly, our model achieves better performance across the whole M4 hourly dataset than all baseline models, and in particular  $39.4\%$ better than LSTM. Notably, the closest model turned out to be SARIMAX with seasonality period 24, which outperformed other machine learning-based models which are not directly modelled for seasonality unlike a part of SARIMAX. Thereby, this suggests that our model is better at detecting seasonality patterns, which are dominantly seen in this particular dataset. We note that no preprocessing specific to seasonality extraction has been made. Similar to the real life datasets, the Naive model has the highest error metric in all datasets, and the disjoint model performs consistently worse than our joint model. LightGBM and MLP performs comparably better than our baseline models, though they have higher errors in hourly and yearly datasets where seasonality and sparsity were emphasized, respectively. Overall, the hybrid model achieves better than both its disjoint counterpart and various standalone models including LSTM on average over three subset of M4 competition datasets of varying characteristics.
\begin{table}[!t]
    \centering
    \begin{tabular}{c c c c} 
                   &         Hourly   &          Daily   &          Yearly \\
        \hline
        Naive      &         0.50915  &         0.09608  &         0.67152  \\ 
        SARIMAX    &         0.15549  &         0.05425  &         0.20032  \\ 
        MLP        &         0.19380  &         0.02940  &         0.22447  \\ 
        LightGBM   &         0.19468  & \textbf{0.01602} &         0.38881\\ 
        LSTM       &         0.20282  &         0.06448  &         0.18032  \\ 
        Disjoint   &         0.37580  &         0.07860  &         0.56895  \\ 
        LSTM-SGBDT & \textbf{0.12290} &         0.01658 & \textbf{0.15852} \\
        \hline
    \end{tabular}
    \caption{Mean MAPE performances of the models across M4 datasets.}
    \label{table:m4}
\end{table}

\subsubsection{Case Study on a Hourly Time Series}
\begin{figure}[!t]
\centering
\includegraphics[width=\linewidth]{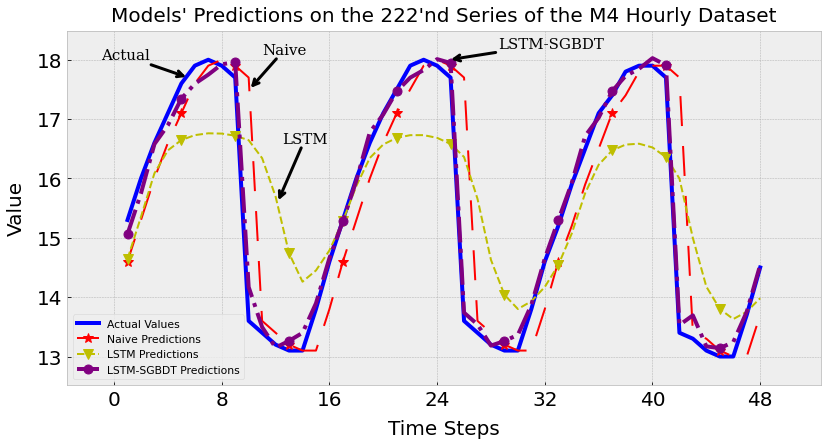}
\caption{Naive, LSTM and LSTM-SGBDT predictions over the forecast horizon of an M4 hourly series.}
\label{fig:m4_hourly_example}
\end{figure}

As a case study to show the effectiveness of our proposed model visually, we look at the testing part of an example series (222nd) from the hourly dataset of length 48 along with the predictions of LSTM, LSTM-SGBDT and the naive model. The plot of predictions are given in Fig. \ref{fig:m4_hourly_example}. The corresponding MAPE metrics of the naive, LSTM and LSTM-SGBDT models are 0.0417, 0.0723 and 0.0096, respectively. We observe that the vanilla LSTM model failed to capture the amplitude of the seasonal pattern. It also tends to predict the last observed value (i.e., the same as the naive predictions) for many time steps, which LSTM-based models are known to suffer from \cite{lstm_naive}. These lead to LSTM being a poorer predictor than the naive model in this specific time series. Our model, on the other hand, successfully captures the seasonal pattern and closely follows the actual values. 

\subsubsection{Recursive Prediction over the Hourly Dataset}
In this section, we perform recursive prediction with our model. To this end, we use the entire hourly dataset that M4 competition provided \cite{m4_comp}. There are in total 414 time series in this dataset, of which 245 are of length 1008 and the remaining 169 are of length 748. The competition demanded a 48-step prediction without revealing the true values as the prediction arrives, which distinguishes this task from the one-step ahead prediction we performed in the previous section. We use the last 48 samples of each series as a hold-out set, over which we validate our model. We note that the validation was performed for each series separately. We extend one-step ahead forecasting to this scenario as follows. We first predict the first sample of the test set. Then, treating this prediction as if it was the true value, we predict the second sample of the test set without refitting. This procedure recursively continues until we produce all of the 48 predictions. As for the model configuration, we use 1-layer LSTM network in the front end and use last pooling in its last layer. The tuned hyperparameters are number of hidden neurons for the LSTM network; number of trees, depth of each tree and shrinkage rate for the sGBDT. We also tune the number of epochs and the learning rate of the gradient-based updates. We use the same preprocessing procedure as described in the previous section. For the performance metric, to comply with the M4 competition results, we use symmetric mean absolute percentage error (sMAPE), which is given as
\begin{equation*}
    \text{sMAPE}({\boldsymbol{d}}_H, {\boldsymbol{\hat d}}_H) = \frac{2}{H}\sum_{i=1}^{H}\, \frac{|{d_{H,i} - \hat{d}_{H,i}}|}{|{d_{H,i}|} + |{\hat{d}_{H,i}}|},
\end{equation*}
where $H$ is the prediction horizon (e.g, 48 hours), ${\boldsymbol{d}}_H$ and ${\boldsymbol{\hat d}}_H$ denote the vector of true and predicted values over the horizon, respectively.

There are 29 sMAPE results for the hourly dataset reported in the M4 competition \cite{m4_comp}, some of which are benchmark results, e.g., simple exponential smoothing. For brevity, we do not include the results that were ranked after (and including) the best benchmark result, which was the multi-layer perceptron benchmark. Therefore, 12 results including ours are reported in Table \ref{table:m4_hourly_recursive} in sorted order.
\begin{table}[!b]
    \centering
    \begin{adjustbox}{width=1\textwidth}
    \begin{tabular}{c c c c c} 
         & Doornik et al. &   Smyl & Pawlikowski et al. & Pedregal et al. \\
        \hline
        &           8.913 &  9.328 &               9.611 &            9.765\\[4pt]
        & Jaganathan \& Prakash & Montero-Manso et al. & Darin \& Stellwagen & LSTM-SGBDT\\
        \hline
        &                9.934 &                11.506 &             11.683 &     11.871\\[4pt]
        & Waheeb & Roubinchtein & Petropoulos \& Svetunkov &   Shaub\\
        \hline
        &  12.047 &       12.871 &                  13.167 &  13.466\\
    \end{tabular}
    \end{adjustbox}
    \caption{Mean sMAPE (\%) metrics of the selected 12 models for the recursive estimation experiment over 414 hourly time series from the M4 dataset.}
    \label{table:m4_hourly_recursive}
\end{table}

We note that the contestants had come up with models that were tailored for the dataset, performed extensive feature engineering, and also tuned data preprocessing strategies during the competition. Our model, on the other hand, employs no special feature engineering and only applies standardization as the preprocessing step (as done in all of the experiments in Section 3). We only used the past values of the target signal for each series, and due to its generic nature, the LSTM network on the front end of the model is tasked to generate relevant features from this raw sequential data and feed it to the supervised regressor sGBDT on the back end. We observe that our model still achieves a comparable performance among contestants when using the recursive estimation strategy over this dataset.

\subsection{Financial Datasets}\label{sec:fin_datasets}
In this section, we show the generic nature of our framework. To this end, we replace the LSTM network in the front end of the model with a gated recurrent unit (GRU) network \cite{gru_paper}. A GRU cell, similar to an LSTM cell, is endowed with gated routing mechanisms in order to avoid the vanishing gradient problem \cite{rnn_vanishes}. In particular, the forward pass equations in one cell are described as
\begin{equation}\label{eq:gruforward}
\begin{aligned}
    \boldsymbol{z}_t &= \sigma(\boldsymbol{W}_z \,\{\boldsymbol h_{t-1}, \boldsymbol x_t\} + \boldsymbol b_z)\\
    \boldsymbol{r}_t &= \sigma(\boldsymbol{W}_r \,\{\boldsymbol h_{t-1}, \boldsymbol x_t\} + \boldsymbol b_r)\\
    \tilde{\boldsymbol h}_t &= \tanh(\boldsymbol{W}_h \,\{\boldsymbol{r}_t \odot \boldsymbol h_{t-1}, \boldsymbol x_t\} + \boldsymbol b_h)\\
    \boldsymbol{h}_t &= (1 - \boldsymbol{z}_t) \odot \boldsymbol{h}_{t-1} + \boldsymbol{z}_t \odot \tilde{\boldsymbol h}_t,
\end{aligned}
\end{equation}
where $\boldsymbol h_t$ is the hidden state vector, $\boldsymbol x_t$ is the input to the cell, $\boldsymbol h_{t-1}$ is the hidden state from the previous cell and $\{\boldsymbol h_{t-1}, \boldsymbol x_t\}$ denotes the vertical concatenation of $\boldsymbol x_t$ and $\boldsymbol h_{t-1}$. $\boldsymbol z_t$ and $\boldsymbol r_t$ are the update and reset gate vectors, respectively, and $\tilde{\boldsymbol h}_t$ is the candidate state vector, whose corresponding stacked weight matrices are $\boldsymbol W_z, \boldsymbol W_r$ and $\boldsymbol W_h$. The corresponding biases are represented by $\boldsymbol b_z, \boldsymbol b_r$ and $\boldsymbol b_h$. $\sigma$ denotes the logistic sigmoid function and $\tanh$ is the hyperbolic tangent function; both apply point-wise. To implement a GRU-SGBDT, we directly replace the LSTM equations with GRU equations; this shows the flexibility of the framework for the sequential feature extraction.
\begin{figure}[!b]
\centering
\includegraphics[width=\linewidth]{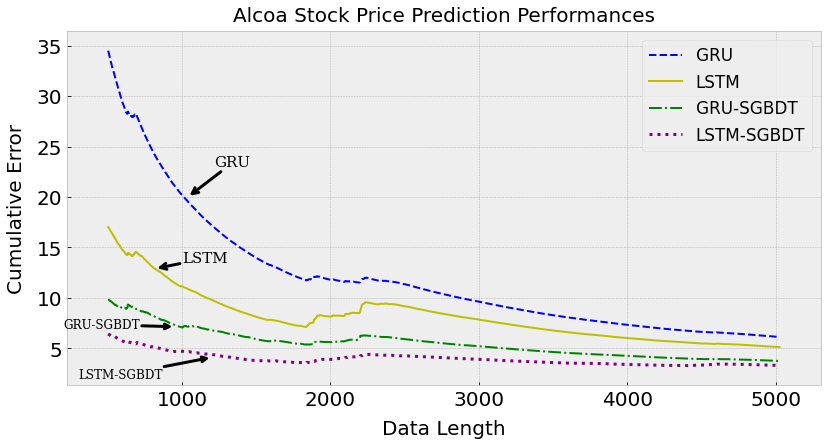}
\caption{Time accumulated errors of standalone and hybrid models for the Alcoa stock price dataset for the period 2000-2020.}
\label{fig:alcoa_perf}
\end{figure}
We consider four models in this section: LSTM, GRU, LSTM-SGBDT and GRU-SGBDT. We evaluate the performances of the models under two financial scenarios in an online setting. We first consider the Alcoa stock price dataset \cite{alcoa_dataset}, which contains the daily stock price values between the years 2000 and 2020. We aim to predict the future prices values by considering the past prices. We only apply standardization to the price values in a cumulative manner. For standardization, we update the stored mean and standard deviation as more data are revealed. We hold out $10\%$ of the data from the beginning and use it for the hyperparameter validation. We choose a window size of 5 so that the last five trading days' stock prices are used by the models to predict the today's price value. Tuned hyperparameters are the same as those in M4 datasets. Fig. \ref{fig:alcoa_perf} illustrates the performance of the models as the data length varies. We observe that hybrid models consistently achieve lower accumulated errors than the standalone recurrent models, showing the benefits of the proposed framework. GRU network starts off with a huge error margin but closes the gap after ~3000 data points, although it does not surpass LSTM overall. LSTM-SGBDT performs closer to GRU-SGBDT yet better in each time step, though the gap is essentially closed when all data are revealed. In addition, LSTM and GRU models are rather slower to react to the abrupt change in the stock prices around length of 2,000 which corresponds to the global financial crisis in the late 2008.

\begin{figure}[!t]
\centering
\includegraphics[width=\linewidth]{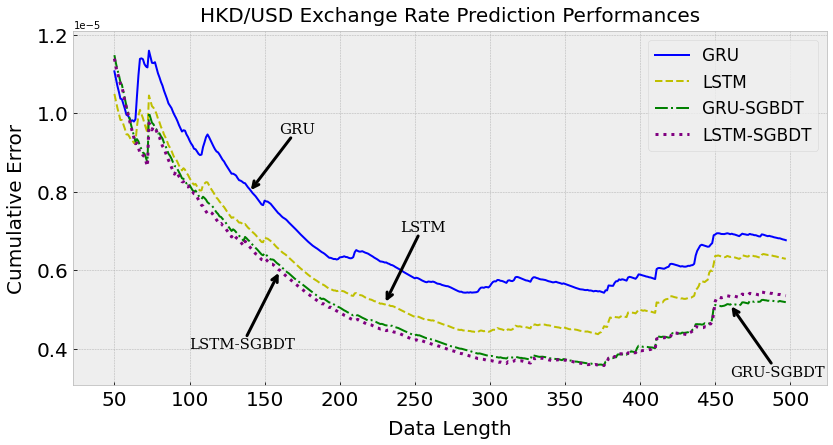}
\caption{Time accumulated errors of standalone and hybrid models for the Hong Kong exchange rate against U.S. dollars.}
\label{fig:hk_ex_perf}
\end{figure}

Apart from the Alcoa stock price dataset, we also experiment with the Hong Kong exchange rate dataset \cite{hk_dataset}, which has the daily rate for Hong Kong dollars against US dollars between the dates 2005/04/01 and 2007/03/30. Our goal is to predict the future exchange rates by using the data of the previous five days. We use the same setup for the hyperparameter configuration as in the Alcoa dataset. The time accumulated errors for both models are presented in Fig. \ref{fig:hk_ex_perf}. We observe that while all models follow a similar error pattern, our hybrid models again consistently achieve a lower cumulative error than the standalone models for almost all time steps. The cumulative errors of the two hybrid models follow each other quite closely, although GRU-SGBDT has a slightly better final accumulated error than LSTM-SGBDT. We also notice that the gap between the errors of standalone models and the hybrid models becomes wider as more data are revealed.
\section{Conclusions}\label{sec:Conclusion}
We studied nonlinear prediction/regression in an online setting and introduced a hybrid architecture composed of an LSTM and a soft GBDT. The recurrent network in the front end acts as a feature extractor from the raw sequential data and the boosted trees in the back end employ a supervised regressor role. We thereby remove the need for hand-designed features for the boosting tree while enjoying joint optimization for the end-to-end model thanks to the soft decision trees as the weak learners in the boosting chain. We derive the gradient updates to be used in backpropagation for all the parameters and also empirically verified the integrity of the architecture. We note that our framework is generic so that one can use other deep learning architectures for feature extraction (such as RNNs and GRUs) and machine learning algorithms for decision making as long as they are differentiable. We achieve consistent and significant performance gains in our experiments over conventional methods on various well-known real life datasets. We also provide the source code for replicability.


\bibliographystyle{elsarticle-num} 
\bibliography{refs}

\end{document}